%% file: main.tex
\documentclass[twoside]{article}

\usepackage[accepted]{aistats2018}

\usepackage[utf8]{inputenc} 
\usepackage[T1]{fontenc}    
\usepackage{hyperref}       
\usepackage{url}            
\usepackage{booktabs}       
\usepackage{amsfonts}       
\usepackage{nicefrac}       
\usepackage{microtype}      
\usepackage{amsmath}
\usepackage{algorithm}
\usepackage{algorithmic}
\usepackage{amsthm}
\usepackage{amssymb}
\usepackage{tikz}
\usetikzlibrary {positioning}
\usepackage{graphicx}
\usepackage{subcaption}
\usepackage{mathtools}
\usepackage[T1]{fontenc}
\usepackage{lmodern}
\usepackage{qtree}
\usepackage[square]{natbib}
\usepackage{sidecap}
\usepackage[toc,page]{appendix}

\newtheorem{prop}{Property}

\newtheorem{proposition}{Proposition}

\mathchardef\mhyphen="2D 
\usepackage{color}
\usepackage{float}
\usepackage{pdfpages}

\usepackage{titlesec}

\DeclareMathOperator*{\argmin}{argmin}
\newcommand{\mysection}[1]{\vspace{-2mm}\section{#1}\vspace{-2mm}}
\newcommand{\mysubsection}[1]{\vspace{-1mm}\subsection{#1}\vspace{-1mm}}

\newcommand{\myparagraph}[1]{\vspace{-2mm}\paragraph{#1}}
\newcommand{\mycaption}[1]{\vspace{0mm}\caption{#1}\vspace{0mm}}
\begin{document}


\twocolumn[

\aistatstitle{Worst-case Optimal Submodular Extensions for Marginal Estimation}

\aistatsauthor{Pankaj Pansari$^{1, 2}$ \And Chris Russell $^{2, 3}$\And M. Pawan Kumar$^{1, 2}$}

\aistatsaddress{ 1: University of Oxford\\
  \texttt{pankaj@robots.ox.ac.uk} \And 2: The Alan Turing Institute \\  \texttt{crussell@turing.ac.uk}
 \And 3: University of Surrey \\ \texttt{pawan@robots.ox.ac.uk}}  
]


\input{./abstract}

\input{./paper}


\bibliographystyle{apalike}
\bibliography{biblio}

\onecolumn
\setcounter{section}{0}
\input{./appendix_arxiv}

\end{document}

%% file: abstract.tex
\begin{abstract}
    Submodular extensions of an energy function can be used to efficiently
    compute approximate marginals via variational inference. The accuracy of
    the marginals depends crucially on the quality of the submodular extension.
    To identify the best possible extension, we show an equivalence between the
    submodular extensions of the energy and the objective functions of linear
    programming (LP) relaxations for the corresponding MAP estimation problem.
    This allows us to (i) establish the worst-case optimality of the submodular
    extension for Potts model used in the literature; (ii) identify the
    worst-case optimal submodular extension for the more general class of
    metric labeling; and (iii) efficiently compute the marginals for the widely
    used dense CRF model with the help of a recently proposed Gaussian filtering method. Using synthetic and real data, we show that our approach provides comparable upper bounds on the log-partition function to those obtained using tree-reweighted message passing (TRW) in cases where the latter is computationally feasible. Importantly, unlike TRW, our approach provides the first practical algorithm to compute an upper bound on the dense CRF model.
\end{abstract}

%% file: paper.tex
\mysection{Introduction}
The desirable optimization properties of submodular set functions have been
widely exploited in the design of approximate MAP estimation algorithms for
discrete conditional random fields (CRFs) \citep{boykov2001fast,
kumar2011improved}. Submodularity has also been recently used to design an
elegant variational inference algorithm to compute the marginals of a discrete
CRF by minimising an upper-bound on the log-partition function. In the initial
work of \citep{djolonga2014map}, the energy of the CRF was restricted to be submodular. In a later work ~\citep{zhang2015higher}, the algorithm was extended to handle more general Potts energy functions. The key idea here is to define a large ground set such that its subsets represent valid labelings, sublabelings or even incorrect labelings (these may assign two separate labels to a random variable and hence be invalid). Given the large ground set, it is possible to define a submodular set function whose value is equal to the energy of the CRF for subsets that specify a valid labeling of the model. We refer to such a set function as a {\em submodular extension} of the energy. 

For a given energy function, there exists a large number of possible submodular extensions. The accuracy of the variational inference algorithm depends crucially on the choice of the submodular extension. Yet, previous work has largely ignored the question of identifying the best extension. Indeed, the difficulty of identifying submodular extensions of general energy functions could be a major reason why the experiments of~\citep{zhang2015higher} were restricted to the special case of models specified by the Potts energy functions.

In this work, we establish a hitherto unknown connection between the submodular
extension of the Potts model proposed by \citet{zhang2015higher}, and the
objective function of an accurate linear programming (LP) relaxation of the
corresponding MAP estimation problem~\citep{kleinberg2002approximation}. This
connection has three important practical consequences. First, it establishes
the accuracy of the submodular extension of the Potts model, via the
UGC-hardness worst-case optimality of the LP relaxation. Second, it provides an
accurate submodular extension of the hierarchical Potts model, via the LP
relaxation of the corresponding MAP estimation problem proposed
by~\citet{kleinberg2002approximation}. Since any metric can be accurately
approximated as a mixture of hierarchical Potts
models~\citep{bartal1996probabilistic, bartal1998approximating}, this result
also provides a computationally feasible algorithm for estimating the marginals
for metric labeling. Third, it establishes the equivalence between the
subgradient of the LP relaxation and the conditional gradient of the problem of
minimising the upper bound of the log-partition. This allows us to employ the
widely used dense CRF, since the subgradient of its LP relaxation can be
efficiently computed using a recently proposed modified Gaussian filtering
algorithm~\citep{ajanthan2017efficient}. As a consequence, we provide the first
efficient algorithm to compute an upper bound of the log-partition function of
dense CRFs. This provides complementary information to the popular mean-field
inference algorithm for dense CRFs, which computes a lower bound on the
log-partition ~\citep{koltun2011efficient}. We show that the quality of our
solution is comparable to tree reweighted message passing (TRW)
~\citep{wainwright2005new} for the case of sparse CRFs. Unlike our approach,
TRW is computationally infeasible for dense CRFs, thereby limiting its use in
practice. Using dense CRF models, we perform stereo matching on standard
data sets and obtain better results than \citep{koltun2011efficient}. The complete code is available at \url{https://github.com/pankajpansari/denseCRF}.

\mysection{Preliminaries}
\label{sec:prelim}

We now introduce the notation and definitions that we will make use of in the remainder of the paper.

\myparagraph{Submodular Functions} Given a ground set $U = \left\{1, \dots, N \right\}$, denote by $2^U$ its power set. A set function $F: 2^U \to \mathbb{R}$ is {\it submodular} if, for all subsets $A, B \subseteq U$, we have
\begin{equation}
    F(A \cup B) + F(A \cap B) \leq F(A) + F(B).
\end{equation}
The set function $F$ is {\it modular} if there exists ${\bf s} \in \mathbb{R}^N$ such that $F(A) = \sum_{k \in A} s_k \ \forall \ A \subseteq 2^U$. Henceforth, we will use the shorthand $s(A)$ to denote ${\sum_{k \in A}} s_k$.

\myparagraph{Extended Polymatroid} Associated with any submodular function $F$ is a special polytope known as the {\it extended polymatroid} defined as
\begin{align}
EP(F) &= \{{\bf s} \in \mathbb{R}^N | \enskip \forall A \subseteq U: s(A) \leq F(A)\},
\label{eq:epf}
\end{align}
where $\bf s$ denotes the modular function $s(.)$ considered as a vector. 

\myparagraph{Lovasz Extension} For a given set function $F$ with $F(\emptyset) = 0$, the value of its Lovasz extension $f({\bf w}): \mathbb{R}^N \to \mathbb{R}$ is defined as follows: order the components of $\bf w$ in decreasing order such that $w_{j_1} \geq w_{j_2} \geq \dots \geq w_{j_N}$, where $(j_1, j_2, \dots, j_N)$ is the corresponding permutation of the indices. Then,
\begin{align}
    f({\bf w}) = \sum_{k = 1}^{N} w_{j_k} \left(F(j_1, \dots, j_k) - F(j_1,
    \dots, j_{k - 1})\right).
    \label{eq:lovasz_def}
\end{align}
The function $f$ is an extension because it equals $F$ on the vertices of the
unit cube. That is, for any $A \subseteq V$, $f({\bf 1}_A) = F(A)$ where ${\bf
1}_A$ is the $0\mhyphen{1}$ indicator vector corresponding to the elements of $A$.

{\prop By Edmond's greedy algorithm \citep{edmonds1970submodular}, if ${\bf w} \geq 0$ (non-negative elements), 
\begin{equation}
    f({\bf w}) = \max_{{\bf s} \in EP(F)} \langle {\bf w}, {\bf s} \rangle.
    \label{eq:lovasz_max}
\end{equation}
\label{prop:greedy}}

Property \ref{prop:greedy} implies that an LP over $EP(F)$ can be solved by computing the value of the Lovasz extension using equation (\ref{eq:lovasz_def}).

{\prop The Lovasz extension $f$ of a submodular function $F$ is a convex piecewise linear function.\label{prop:lovasz_convex}}

Property \ref{prop:lovasz_convex} holds since $f({\bf w})$ is the pointwise maximum of linear functions according to equation (\ref{eq:lovasz_max}).

\myparagraph{CRF and Energy Functions}
A CRF is defined as a graph on a set of random variables ${\mathcal X} = \{X_1,\dots, X_N\}$ related by a set of edges $\mathcal N$. We wish to assign every variable $X_a$ one of the labels from the set $\mathcal{L} = \{1, 2, \dots, L\}$. The quality of a labeling $\bf x$ is given by an energy function defined as 
\begin{equation}
    E({\bf x}) = \sum_{a \in {\mathcal X}} \phi_a(x_a) + \sum_{(a, b) \in {\mathcal N}} \phi_{ab}(x_a, x_b),
    \label{eq:energy_min}
  \end{equation}
where $\phi_a$ and $\phi_{ab}$ are the unary and pairwise potentials respectively. In computer vision, we often think of $\mathcal X$ as arranged on a grid. A \emph{sparse CRF} has $\mathcal N$ defined by 4-connected or 8-connected neighbourhood relationships. In a \emph{dense CRF}, on the other hand, every variable is connected to every other variable.

The energy function also defines a probability distribution $P({\bf x})$ as follows:
\begin{equation}
    P({\bf x}) =
    \begin{cases}
        \frac{1}{\mathcal{Z}} \exp(-E({\bf x})) & \quad \text{if} \enskip {\bf x} \in {\mathcal L}^N,\\
       0  & \quad \text{otherwise}. \\
    \end{cases}
\label{eq:crf_prob_distrib}
\end{equation}
The normalization factor $Z = \sum_{{\bf x} \in {\mathcal L}^N} \exp(-E({\bf x}))$ is known as the {\it partition function}. 

\myparagraph{Inference} There are two types of inference problems in CRFs: 

(i) Marginal inference: We want to compute the marginal probabilities $P(X_a = i)$ for every $a = 1, 2, \dots, N$ and $i = 1, 2, \dots, L$.

(ii) MAP inference: We want to find a labeling with the minimum energy, that is, $\min_{{\bf x} \in {\mathcal L}^N} E({\bf x})$. Equivalently, MAP inference finds the mode of $P({\bf x})$.

\mysection{Review: Variational Inference Using Submodular Extensions}
\label{sec:review}
We now summarise the marginal inference method of \citet{zhang2015higher}. To do this, we need to first define submodular extensions.

\begin{figure}
\centering
\includegraphics[scale = 0.30]{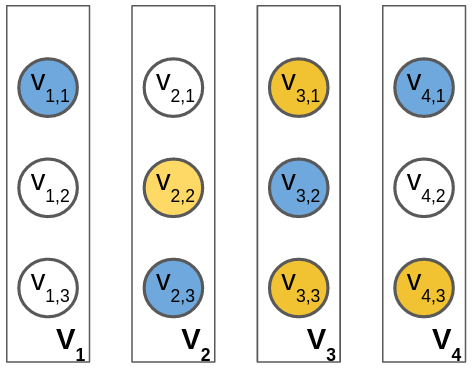}
\mycaption{\footnotesize \em Illustration of $1$-of-$L$ encoding used in \citep{zhang2015higher} with 4 variables and 3 labels. The blue labeling, corresponding to $X_1 = 1, X_2 = 3, X_3 = 2, X_4 = 1$, is valid. The yellow labeling, corresponding to $X_2 = 2, X_3 = 1, 3, X_4 = 3$, is invalid since $X_3$ has been assigned multiple labels and $X_1$ has been assigned none.} 
\label{fig:encoding}
\end{figure}

\myparagraph{Submodular Extensions} A submodular extension is defined using
a ground set such that some of its subsets correspond to valid CRF labelings.
To such an extension, we need an encoding scheme which gives the sets corresponding to valid CRF labelings. 

One example of an encoding scheme is the $1$-of-$L$ encoding, illustrated in figure \ref{fig:encoding}. Let each variable $X_a$ take one of $L$ possible labels. In this scheme, we represent the set of possible assignments for $X_a$ by the set $V_a = \{v_{a1}, v_{a2}, \dots, v_{aL}\}$. If $X_a$ is assigned label $i$, then we select the element $v_{ai}$. Extending to all variables, our ground set becomes $V = \cup_{a = 1}^N V_a$. A valid assignment $A \subseteq V$ assigns each variable exactly one label, that is, $|A \cap V_a| = 1$ for all $V_a$. We denote the set of valid assignments by $\mathcal M$ where $\mathcal{M} = \cap_{a = 1}^{N} \mathcal{M}_a$ and $\mathcal{M}_a = \{A: |A \cap V_a| = 1\}$.

Using our ground set $V$, we can define a submodular function $F$ which equals $E(\bf x)$ for all sets corresponding to valid labelings, that is, $F(A_{\bf x}) = E({\bf x}),\ {\bf x} \in {\mathcal L}^N$ where $A_{\bf x}$ is the set encoding of $\bf x$. We call such a function $F$ a {\it submodular extension} of $E({\bf x})$. 

\myparagraph{Upper-Bound on Log-Partition} Using a submodular extension $F$ and given any ${\bf s} \in EP(F)$, we can obtain an upper-bound on the partition function as
\begin{equation}
    {\mathcal Z} = \sum_{A \in {\mathcal M}} \exp(-F(A)) \leq \sum_{A \in {\mathcal M}} \exp(-s(A)),
\end{equation}
where $\mathcal M$ is the set of valid labelings. The upper-bound is the partition function of the distribution $Q(A) \propto \exp(-s(A))$, which factorises fully because $s(.)$ is modular. Since ${\bf s} \in EP(F)$ is a free parameter, we can obtain good approximate marginals of the distribution $P(\cdot)$ by minimising the upper-bound. After taking logs, we can equivalently write our optimisation problem as
\begin{equation}
  \min_{{\bf s} \in EP(F)} g({\bf s}), \text{where } g({\bf s}) = \log \sum_{A \in {\mathcal M}} \exp(-s(A)).
\label{eq:min_upper_bound}
\end{equation}

\myparagraph{Conditional Gradient Algorithm} The conditional gradient algorithm
(algorithm \ref{algo:fw_inference}) \citep{frank1956algorithm} is a good
candidate for solving problem (\ref{eq:min_upper_bound}) due to two reasons.
First, problem \eqref{eq:min_upper_bound} is convex. Second, as solving an LP over EP(F) is computationally tractable (property \ref{prop:greedy}), the conditional gradient can be found efficiently. The algorithm starts with an initial solution $\bf s_0$ (line \ref{line:initial}). At each iteration, we compute the conditional gradient $\bf s^*$ (line \ref{line:cond_grad}), which minimises the linear approximation $g({\bf s}_k) + \nabla g({\bf s}_k)^T ({\bf s} - {\bf s}_k)$ of the objective function. Finally, $\bf s$ is updated by either (i) fixed step size schedule, as in line \ref{line:step_size} of algorithm \ref{algo:fw_inference}, or (ii) by doing line search ${\bf s}_{k+1} = \min_{0 \leq \gamma \leq 1} g(\gamma {\bf s}^* + (1 - \gamma) {\bf s}_{k})$.

\begin{algorithm}
\mycaption{Upper Bound Minimisation using Conditional Gradient Descent}
\label{algo:fw_inference}
\begin{algorithmic}[1]
    \STATE Initialize ${\bf s} = {\bf s}_0 \in EP(F)$ \label{line:initial}
\FOR{k = 1 to \texttt{MAX\_ITER}}
\STATE ${\bf s^*} = \argmin_{{\bf s} \in EP(F)} \langle \nabla g({\bf s}_k), {\bf s} \rangle$ \label{line:cond_grad}
\IF{$\langle {\bf s^* - s}_k, \nabla g({\bf s}_k) \rangle \leq \epsilon$}
\STATE {\bf break}
\ENDIF
\STATE ${\bf s}_{k + 1} = {\bf s}_k + \gamma({\bf s^*} - {\bf s}_k)$ \text{with} $\gamma = 2/(k + 2)$ \label{line:step_size}
\ENDFOR   
\RETURN $\bf s$
\end{algorithmic}
\end{algorithm}

\section{Worst-case Optimal Submodular Extensions via LP Relaxations}

\myparagraph{Worst-case Optimal Submodular Extensions} Different choices of extensions $F$ change the domain in problem \eqref{eq:min_upper_bound}, leading to different upper bounds on the log-partition function. How does one come up with an extension which yields the tightest bound? 

In this paper, we focus on submodular extension families $\mathcal F(.)$ which
for each instance of the energy function $E(.)$ belonging to a given class $\mathcal E$ gives a corresponding
submodular extension ${\mathcal F}(E)$. We find the extension family ${\mathcal
F}_{opt}$ that is \emph{worst-case optimal}. This implies that there does not
exist another submodular extension family ${\mathcal F}$ that gives a tighter
upper bound for problem \eqref{eq:min_upper_bound} than ${\mathcal F}_{opt}$
for all instances of the energy function in $\mathcal E$. Formally, 
\begin{equation}
    \nexists {\mathcal F}: \min_{{\bf s} \in EP({\mathcal F}(E))} g({\bf s})
    \leq \min_{{\bf s} \in EP({\mathcal F}_{opt}(E))} g({\bf s}) \enspace
    \forall \; E(.) \in \mathcal E.
    \label{eq:best_submod}
\end{equation}
Note that our problem is different from taking a given energy model and
obtaining a submodular extension which is optimal for that model. Also, we seek
a closed-form analytical expression for $\mathcal F$. For the sake of clarity,
in the analysis that follows we use $F$ to represent ${\mathcal F}(E)$ where
the meaning is clear from context. The two classes of energy functions we
consider in this paper are Potts and hierarchical Potts families.
\myparagraph{Using LP Relaxations} If we introduce a temperature parameter in $P({\bf x})$ (equation \eqref{eq:crf_prob_distrib}) by using $E({\bf x})/T$ and decrease $T$, the resulting distribution starts to peak more sharply around its mode. As $T \to 0$, marginal estimation becomes the same as MAP inference since the resulting distribution $P^0({\bf x})$ has mass 1 at its mode ${\bf x}^*$ and is 0 everywhere else. Given the MAP solution ${\bf x}^*$, one can compute the marginals as $P(X_i = j) = [x_i^* = j]$, where [.] is the Iverson bracket.

Motivated by this connection, we ask if one can introduce a temperature parameter to our problem \eqref{eq:min_upper_bound} and transform it to an LP relaxation in the limit $T \to 0$? We can then hope to use the tightest LP relaxations of MAP problems known in literature to find worst-case optimal submodular extensions. We answer this question in affirmative. Specifically, in the following two sections we show how one can select the set encoding and submodular extension to convert problem \eqref{eq:min_upper_bound} to the tightest known LP relaxations for Potts and hierarchical Potts models. Importantly, we prove the worst-case optimality of the extensions thus obtained. 

\begin{figure}
\centering
\includegraphics[scale = 0.30]{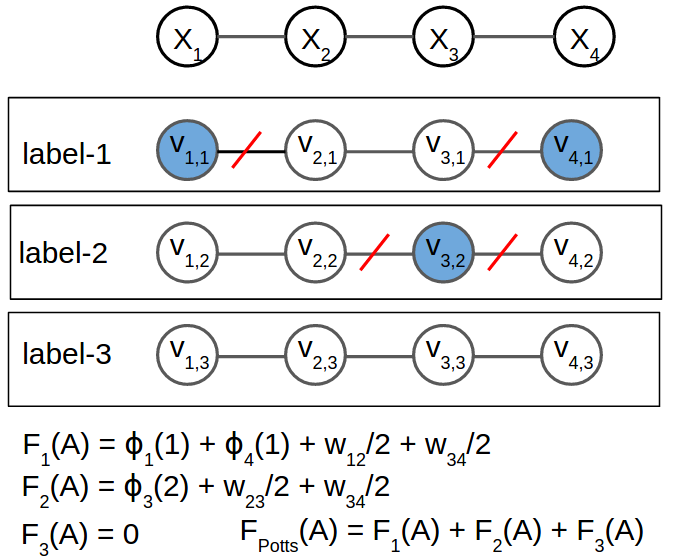}
\mycaption{\footnotesize \em An illustration of the worst-case optimal
submodular extension for Potts model for a chain graph of 4 variables, each of
which can take 3 labels. The figure shows the way to compute the extension
values of the set $A = \{v_{1, 1}, v_{4, 1}, v_{3, 2}$\}.}
\label{fig:alternate}
\end{figure}

\mysection{Potts Model}

The Potts model, also known as the uniform metric, specifies the pairwise potentials $\phi_{ab}(x_a, x_b)$ in equation (\ref{eq:energy_min}) as follows:
\begin{equation}
    \phi_{ab}(x_a, x_b) = w_{ab} \cdot [x_a \neq x_b],
\end{equation}
where $w_{ab}$ is the weight associated with edge $(a, b)$. 
\myparagraph{Tightest LP Relaxation} Before describing our set encoding and submodular extension, we briefly outline the LP relaxation of the corresponding MAP estimation problem. To this end, we define indicator variables $y_{ai}$ which equal 1 if $X_a = i$, and 0 otherwise. The following LP relaxation is the tightest known for Potts model in the worst-case, assuming the Unique Games Conjecture to be true \citep{manokaran2008sdp}
\begin{align}
    \text{(P-LP)} \quad \min_{\bf y} \enskip & E({\bf y}) =  \sum_{a} \sum_{i} \phi_a(i)y_{ai} + \nonumber\\
     &\sum_{(a, b) \in {\mathcal N}} \sum_{i} \frac{w_{ab}}{2} \cdot |y_{ai} - y_{bi}| \nonumber\\
    \text{s.t} \quad &{\bf y} \in \Delta.
\label{eq:p-lp}
\end{align}
The set $\Delta$ is the union of $N$ probability simplices:
\begin{equation}
    \Delta = \{{\bf y}_a \in \mathbb{R}^L | {\bf y}_a \geq 0  \textrm{ and } \langle {\bf 1}, {\bf y}_a \rangle = 1\},
    \label{eq:delta}
\end{equation}
where $\bf y$ is the vector of all variables and ${\bf y}_a$ is the component
of $\bf y$ corresponding to $X_a$.

\myparagraph{Set Encoding} We choose to use the $1$-of-$L$ encoding for Potts model as described in section \ref{sec:review}. With the encoding scheme for Potts model above, $g({\bf s})$ can be factorised and problem (\ref{eq:min_upper_bound}) can be rewritten as:

\begin{equation}
\min_{{\bf s} \in EP(F)} \quad \sum_{a = 1}^{N} \log \sum_{i = 1}^L \exp(-s_{ai}).
\label{eq:potts_upper_bound}
\end{equation}
(See Remark 1 in appendix)

\myparagraph{Marginal Estimation with Temperature} We now introduce
a temperature parameter $T > 0$ to problem \eqref{eq:potts_upper_bound} which
divides $E({\bf x})$, or equivalently divides $\bf s$ belonging to $EP(F)$. Also, since $T > 0$, we can multiply the objective by $T$ leaving the problem unchanged. Without changing the solution, we can transform problem \eqref{eq:potts_upper_bound} as follows
\begin{equation}
    \min_{{\bf s} \in EP(F)} \quad g_T ({\bf s}) = \sum_{a = 1}^{N} T \cdot \log \sum_{i = 1}^L \exp(-\frac{s_{ai}}{T}).
\label{eq:potts_temp}
\end{equation}
\myparagraph{Worst-case Optimal Submodular Extension} 
We now connect our marginal estimation problem (\ref{eq:min_upper_bound}) with
LP relaxations using the following proposition.
{\proposition Using the $1$-of-$L$ encoding scheme, in the limit $T \to 0$,
  problem (\ref{eq:potts_temp}) for Potts model becomes:
\begin{align}
    - \min_{{\bf y} \in \Delta} f({\bf y}) 
\end{align}
where $f(.)$ is the Lovasz extension of $F(.)$.
\newline \textup{(Proof in appendix)}
\label{proposition:potts_equiv}}

The above problem is equivalent to an LP relaxation of the corresponding MAP
estimtation problem (see Remark 2 in appendix). We note that $g_T ({\bf s})$ in problem \eqref{eq:potts_temp} becomes the objective function of an LP relaxation in the limit $T \to 0$. We seek to obtain the worst-case optimal submodular extension by making $g_T ({\bf s})$ same as the objective of (P-LP) as $T \to 0$. Since at $T = 1$, problems \eqref{eq:potts_temp} and \eqref{eq:potts_upper_bound} are equivalent, this gives us the worst-case optimal extension for our problem \eqref{eq:potts_upper_bound} as well. 

The question now becomes how to recover the worst-case optimal submodular extension using $E({\bf y})$. The following propositions answers this question.
{\proposition The worst-case optimal  submodular extension for Potts model is given by $F_{Potts}(A) = \sum_{i = 1}^L F_i(A)$, where
\begin{align}
    F_i(A) &= \sum_a \phi_{a}(i) [|A \cap \{v_{ai}\}| = 1] + \nonumber \\
           &\sum_{(a, b) \in {\mathcal N}} \frac{w_{ab}}{2} \cdot [|A \cap \{v_{ai}, v_{bi}\}| = 1]
\end{align}
Also, $E({\bf y})$ in (P-LP) is the Lovasz extension of $F_{Potts}$.
\textup{(Proof in appendix)}
\label{proposition:potts_worst-case optimal}}
\vspace{-1mm}

Proposition \ref{proposition:potts_worst-case optimal} paves the way for us to
identify the worst-case optimal extension for hierarchical Potts model, which
we discuss in the following section.

\begin{figure}
\centering
\resizebox{150pt}{120pt}{
\begin{tikzpicture}[-latex ,auto, node distance =1.5 cm and 1.5cm ,on grid
  , semithick , state/.style ={ circle , color = black , draw,black , text=blue , minimum width =1 cm}, cross/.style={color=red, very thick, dashed}, unary/.style={color=blue, very thick}] 

  \node[state] (n11) at (1, 1) {\Large $1\mhyphen1$};
  \node[state] (n12) at (3, 1) {\Large $1\mhyphen2$};
  \node[state] (n13) at (5, 1) {\Large $1\mhyphen3$};
  \node[state] (n21) at (7, 1) {\Large $2\mhyphen1$};
  \node[state] (n22) at (9, 1) {\Large $2\mhyphen2$};
  \node[state] (n23) at (11, 1) {\Large $2\mhyphen3$};
  \node[state] (source) at (6, 3) {\Large s};
  \node[state] (sink) at (6, -3) {\Large t};

\path[unary] (n11) edge [bend right =25] node [pos=0.20, anchor=west] {\Large$\phi_1(1)$} (sink);
\path[unary] (n12) edge [bend right =25] node [pos=0.20, anchor=west]{\Large$\phi_1(2)$} (sink);
\path[unary] (n13) edge [bend right =25] node [pos=0.20, anchor=west]{\Large$\phi_1(3)$} (sink);
\path[unary] (n21) edge [bend left =25] node [pos=0.20, anchor=west]{\Large$\phi_2(1)$} (sink);
\path[unary] (n22) edge [bend left =25] node [pos=0.20, anchor=west]{\Large$\phi_2(2)$} (sink);
\path[unary] (n23) edge [bend left =25] node [pos=0.20, anchor=west]{\Large$\phi_2(3)$} (sink);
 
\path[cross] (n11) edge [bend left =25] node[above] {} (n21);
\path[cross] (n21) edge [bend left =25] node[below] {} (n11);

\path[cross] (n12) edge [bend left =25] node[above] {} (n22);
\path[cross] (n22) edge [bend left =25] node[below] {} (n12);

\path[cross] (n13) edge [bend left =25] node[above] {} (n23);
\path[cross] (n23) edge [bend left =25] node[below] {} (n13);
\end{tikzpicture}
}
\mycaption{\footnotesize \em An st-graph specifying the worst-case optimal submodular
extension for Potts model for 2 variables with 3 labels each and connected to
each other. There is a node for each variable and each label, that is, for all
elements of the ground set. The nodes have been labeled as `variable-label',
hence node 1-1 represents the element $v_{11}$ and so on. The solid blue arcs
model the unary potentials, and the dotted red arcs represent the pairwise
potentials. Each dotted red arc has weight $w_{12}/2$.}
\label{fig:optimal_extension_graph}
\end{figure}

\mysection{Hierarchical Potts}
\label{sec:rhst_worst-case optimal}

\begin{figure}
\centering
\includegraphics[scale = 0.30]{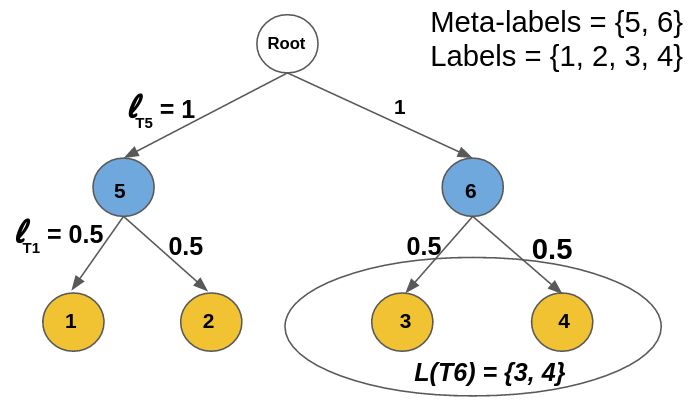}
\mycaption{\footnotesize \em An hierarchical Potts model instance illustrating notations with 2 meta-labels (blue) and 4 labels (yellow). All labels are at the same level. $r = 2$, that is, edge-length decreases by 2 at each level. Also, distance between labels 1 and 3 is $d(1, 3) = 0.5 + 1 + 1 + 0.5 = 3$.} 
\label{fig:rhst}
\end{figure}
Potts model imposes the same penalty for unequal assignment of labels to neighbouring variables, regardless of the label dissimilarity. A more natural approach is to vary the penalty based on how different the labels are. A hierarchical Potts model permits this by specifying the distance between labels using a tree with the following properties:
\vspace{-0.25cm}
\begin{enumerate}
  \itemsep-0.25em 
\item The vertices are of two types: (i) the leaf nodes representing labels, and (ii) the non-leaf nodes, except the root, representing meta-labels.
\item The lengths of all the edges from a parent to its children are the same.
\item The lengths of the edges along any path from the root to a leaf decreases by a factor of at least $r \geq 2$ at each step.
\item The metric distance between nodes of the tree is the sum of the edge lengths on the unique path between them. 
\end{enumerate}
\vspace{-0.25cm}
A subtree $T$ of an hierarchical Potts model is a tree comprising all the descendants of some node $v$ (not root). Given a subtree $T$, $l_T$ denotes the length of the tree-edge leading upward from the root of $T$ and $L(T)$ denotes the set of leaves of $T$. We call the leaves of the tree as labels and all other nodes of the tree expect the root as \emph{meta-labels}. Figure \ref{fig:rhst} illustrates the notations in the context of a hierarchical Potts model.
\myparagraph{Tightest LP Relaxation} We use the same indicator variables $y_{ai}$ that were employed in the LP relaxation of Potts model. Let $y_a(T) = \sum_{i \in L(T)} y_{ai}$. The following LP relaxation is the tightest known for hierarchical Potts model in the worst-case, assuming the Unique Games Conjecture to be true \citep{manokaran2008sdp}
\begin{align}
\text{(T-LP)} \quad \min_{\bf y} \enskip & \widetilde{E}({\bf y}) =  \sum_{a} \sum_{i} \phi_a(i)y_{ai} + \nonumber\\
                                         &\sum_{(a, b) \in {\mathcal N}} w_{ab} \sum_T l_T \cdot |y_{a}(T) - y_{b}(T)| \nonumber\\
    \text{such that} \quad &{\bf y} \in \Delta.
\label{eq:t-lp}
\end{align}
The set $\Delta$ is the same domain as defined in equation (\ref{eq:delta}). We
rewrite this LP relaxation using indicator variables $z_{ai}$ for all labels
and meta-labels as
\begin{align}
    \text{(T-LP-FULL)} &\quad \min \widetilde{E}({\bf z}) \nonumber \\
    {} &\text{such that} \quad {\bf z} \in \Delta'
\label{eq:t-lp-full}
\end{align}
   where $\Delta'$ is the convex hull of the vectors satisfying
\begin{align}
    \enskip &\sum_{i \in \mathcal{L}} z_{ai} = 1, \enskip z_{ai} \in \{0, 1\} \enskip \forall a \in {\mathcal X}, i \in \mathcal{L} \\
    \text{and} \enskip &z_{ai} = \sum_{j \in L(T_i) } z_{aj}.\enskip \forall a \in {\mathcal X}, i \in \mathcal{R - L} \label{eq:consistency_constraint} 
\end{align}
The details of the new relaxation (T-LP-FULL) can be found in the appendix.
\myparagraph{Set Encoding} For any variable $X_a$, let the set of possible assignment of labels and meta-labels be the set $V_a = \{v_{a1}, \dots, v_{aM}\}$, where $M$ is the total number of nodes in the tree except the root. Our ground set is $V = \cup_{a = 1}^{N} V_a$ of size $N \cdot M$.

A consistent labeling of a variable assigns it one label, and all meta-labels on the path from root to the label. Let us represent the set of consistent assignments for $X_a$ by the set $P_a = \{p_{a1}, \dots, p_{aL}\}$, where $p_{ai}$ is the collection of elements from $V_a$ for label $i$ and all meta-labels on the path from root to label $i$. 

The set of valid labelings $A \subseteq V$ assigns each variable exactly one consistent label. This constraint can be formally written as $\mathcal{M} = \cap_{a = 1}^{N} \mathcal{M}_a$ where $\mathcal{M}_a$ has exactly one element from $P_a$.

Let $s'_{ai}$ be the sum of the components of $\bf s$ corresponding to the elements
of $p_{ai}$, that is, 
\begin{equation}
s'_{ai} = \sum_{t \in p_{ai}} s_t.
\label{eq:tree_s}
\end{equation}
Using our encoding scheme, we rewrite problem (\ref{eq:min_upper_bound}) as:
\begin{equation}
\min_{{\bf s} \in EP(F)} \quad \sum_{a = 1}^{N} \log \sum_{i = 1}^L \exp(-s'_{ai}).
\label{eq:metric_upper_bound}
\end{equation}
\myparagraph{Marginal Estimation with Temperature} Similar to Potts model, we now introduce a temperature parameter $T > 0$ to problem \eqref{eq:metric_upper_bound}. The transformed problem becomes

\begin{equation}
    \min_{{\bf s} \in EP(F)} \quad g_T ({\bf s}) = \sum_{a = 1}^{N} T \cdot \log \sum_{i = 1}^L \exp(-\frac{s'_{ai}}{T}).
\label{eq:metric_temp}
\end{equation}

\myparagraph{Worst-case Optimal Submodular Extension} The following proposition connects the marginal estimation problem (\ref{eq:min_upper_bound}) with LP relaxations:

{\proposition In the limit $T \to 0$, problem (\ref{eq:metric_temp}) for hierarchical Potts energies becomes:
\begin{align}
    - \min_{{\bf z} \in \Delta'} f({\bf z}) 
\end{align}
\label{proposition:metric_equiv}}
(Proof in appendix).

The above problem is equivalent to an LP relaxation of the corresponding MAP
estimtation problem (see Remark 3 in appendix). Hence, $g_T ({\bf s})$ becomes the objective function of an LP relaxation in the limit $T \to 0$. We seek to make this objective same as $\widetilde{E}({\bf z})$ of (T-LP-FULL) in the limit $T \to 0$. The question now becomes how to recover the  worst-case optimal submodular extension from $\widetilde{E}({\bf z})$.
{\proposition The  worst-case optimal submodular extension for hierarchical Potts model is given by $F_{hier}(A) = \sum_{i = 1}^M F_i(A)$, where
\begin{align}
    F_i(A) &= \sum_a \phi'_{a}(i) [|A \cap \{v_{ai}\}| = 1] + \nonumber \\
           &\sum_{(a, b) \in {\mathcal N}} {w_{ab}} \cdot l_{T_i} \cdot [|A \cap \{v_{ai}, v_{bi}\}| = 1]
\end{align}
Also, $\widetilde{E}({\bf z})$ in (T-LP-FULL) is the Lovasz extension of $F_{hier}$.
\label{proposition:rhst_worst-case optimal}}
(Proof in appendix)

Since any finite metric space can be probabilistically approximated by mixture
of tree metric \citep{bartal1996probabilistic}, the worst-case optimal
submodular extension for metric energies can be obtained using $F_{hier}$. Note
that $F_{hier}$ reduces to $F_{Potts}$ for Potts model. One can see this by
considering the Potts model as a star-shaped tree with edge weights as 0.5.

\mysection{Fast Conditional Gradient Computation for Dense CRFs}
\label{sec:dense_crf}

\myparagraph{Dense CRF Energy Function} A dense CRF is specified by the following energy function 
\begin{equation}
    E({\bf x}) = \sum_{a \in {\mathcal X}} \phi_a(x_a) + \sum_{a \in {\mathcal X}} \sum_{b \in {\mathcal X}, b \neq a} \phi_{ab}(x_a, x_b).
\end{equation}
Note that every random variable is a neighbour of every other random variable in a dense CRF. Similar to previous work \citep{koltun2011efficient}, we consider the pairwise potentials to be to be Gaussian, that is, 
\begin{align}
    &\phi(i, j) = \mu(i, j) \sum_{m} w^{(m)} k({\bf f}_a^{(m)}, {\bf f}_b^{(m)}), \\
    &k({\bf f}_a^{(m)}, {\bf f}_b^{(m)}) = \exp\left( \frac{- ||{\bf f}_a - {\bf f}_b ||^2}{2} \right).
\end{align}
The term $\mu(i,j)$ is known as {\it label compatibility} function between
labels $i$ and $j$. Potts model and hierarchical Potts models are examples of
$\mu(i, j)$. The other term is a mixture of Gaussian kernels $k(.,.)$ and is
called the {\it pixel compatibility} function. The terms ${\bf f}_a^{(m)}$ are
features that describe the random variable $X_a$. In practice, similar to
\citep{koltun2011efficient}, we use $x, y$ coordinates and RGB values
associated to a pixel as its features.

Algorithm \ref{algo:fw_inference} assumes that the conditional gradient $\bf s^*$ in step {\ref{line:cond_grad}} can be computed efficiently. This is certainly not the case for dense CRFs, since computing $s^*$ involves $NL$ function evaluations of the submodular extension $F$, where $N$ is the number of variables, and $L$ is the number of labels. Each $F$ evaluation has complexity ${\mathcal O}(N)$ using the efficient Gaussian filtering algorithm of \citep{koltun2011efficient}. However, computation of $s^*$ would still be ${\mathcal O} (N^2)$ this way, which is clearly impractical for computer-vision applications where $N \sim 10^5 - 10^6$.

However, using the equivalence of relaxed LP objectives and the Lovasz
extension of submodular extensions in proposition
\ref{proposition:potts_equiv}, we are able to compute $s^*$ in ${\mathcal
O}(NL)$ time. Specifically, we use the algorithm of
\citet{ajanthan2017efficient}, which provides an efficient filtering procedure to compute the subgradient of the LP relaxation objective $E({\bf y})$ of (P-LP).\\
\vspace{-0.5cm}
{\proposition Computing the subgradient of $E({\bf y})$ in (P-LP) is equivalent to computing the conditional gradient for the submodular function $F_{Potts}$. \label{proposition:subgrad}} 
\newline(Proof in appendix).
%
%

A similar observation can be made in case of hierarchical Potts model. Hence we have the first practical algorithm to compute upper bound of log-partition function of a dense CRF for Potts and metric energies.

\mysection{Experiments}
\label{sec:exp}

Using synthetic data, we show that our upper-bound compares favorably with TRW
for both Potts and hierarchical Potts models. For comparison, we restrict
ourselves to sparse CRFs, as the code available for TRW does not scale well to
dense CRFs. We also perform stereo matching using dense CRF models and compare
our results with the mean-field-based approach of \citep{koltun2011efficient}.
All experiments were run on a x86-64, 3.8GHz machine with 16GB RAM. In this
section, we refer to our algorithm as \emph{Submod} and mean field as \emph{MF}.

\begin{figure}
\centering
\begin{tabular}{cc}
\includegraphics[scale = 0.16]{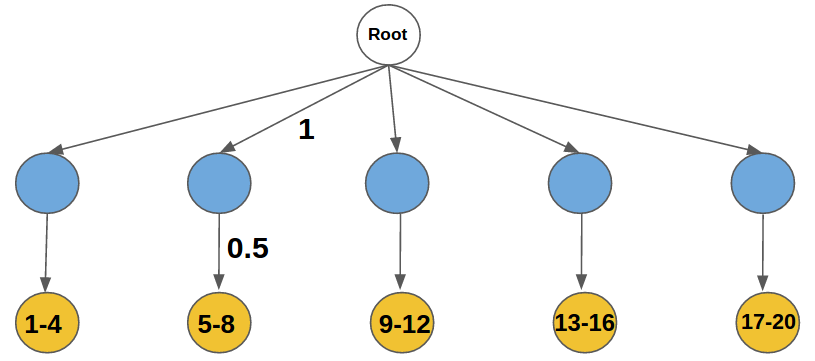} &
\includegraphics[scale = 0.16]{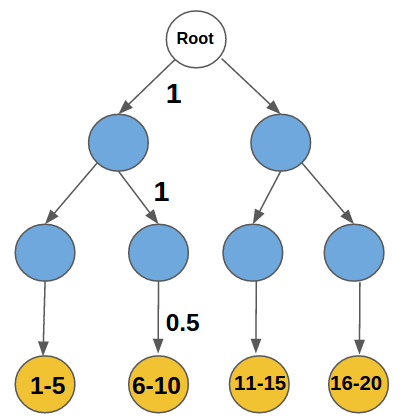} \\
  \scriptsize(a) Tree 1 & \scriptsize(b) Tree 2 \\
\end{tabular}
\mycaption{\footnotesize \em The hierarchical Potts models defining pairwise
distance among 20 labels used for upper-bound comparison with TRW. Blue nodes
are the meta-labels and yellow nodes are labels. All the edges at a particular
level have the same edge weights. The sequence of weights from root level to
leaf level is 1, 0.5 for tree 1 and 1, 1, 0.5 for tree 2. The yellow node is
shown to clump together 4 and 5 leaf nodes for tree 1 and 2 respectively.} 
\label{fig:syn_rhst}
\end{figure}

\begin{figure}
\centering
\resizebox{150pt}{120pt}{
\begin{tikzpicture}[-latex ,auto ,node distance =2 cm and 2cm ,on grid
  , semithick , state/.style ={ circle , color = black , draw,black , text=blue , minimum width =1 cm}, cross/.style={color=red, very thick, dashed}, unary/.style={color=blue, very thick}] 

  \node[state] (n11) at (1, 1) {\Large $1\mhyphen1$};
  \node[state] (n12) at (5, 1) {\Large $1\mhyphen2$};
  \node[state] (n13) at (3, -1) {\Large $1\mhyphen3$};
  \node[state] (n21) at (7, 1) {\Large $2\mhyphen1$};
  \node[state] (n22) at (11, 1) {\Large $2\mhyphen2$};
  \node[state] (n23) at (9, -1) {\Large $2\mhyphen3$};
  \node[state] (source) at (6, 3) {\Large s};
  \node[state] (sink) at (6, -3) {\Large t};
  
  \path[unary] (n11) edge [bend left =25] node[below] {\Large$\phi_1(1)$} (n12);
\path[unary] (n12) edge [bend left =25] node[left] {\Large$\phi_1(2)$} (n13);
\path[unary] (n13) edge [bend left =25] node[left] {\Large$\phi_1(3)$} (n11);
\path[unary] (n21) edge [bend left =25] node[below] {\Large$\phi_2(1)$} (n22);
\path[unary] (n22) edge [bend left =25] node[right] {\Large$\phi_2(2)$} (n23);
\path[unary] (n23) edge [bend left =25] node[right] {\Large$\phi_2(3)$} (n21);

\path[cross] (n11) edge [bend left =25] node[above] {} (n21);
\path[cross] (n21) edge [bend left =25] node[below] {} (n11);

\path[cross] (n12) edge [bend left =25] node[above] {} (n22);
\path[cross] (n22) edge [bend left =25] node[below] {} (n12);

\path[cross] (n13) edge [bend left =25] node[above] {} (n23);
\path[cross] (n23) edge [bend left =25] node[below] {} (n13);
\end{tikzpicture}
}
\mycaption{\footnotesize \em An st-graph specifying the alternate submodular
extension for Potts model for 2 variables with 3 labels each and connected to
each other. The convention used is same as in figure \ref{fig:optimal_extension_graph}. Each dotted red arc has weight $w_{12}/2$.  This alternate extension was also used to derive the extension for hierarchical Potts model.}
\label{fig:alternate_extension}
\end{figure}

\begin{figure}
\centering
\includegraphics[scale = 0.35]{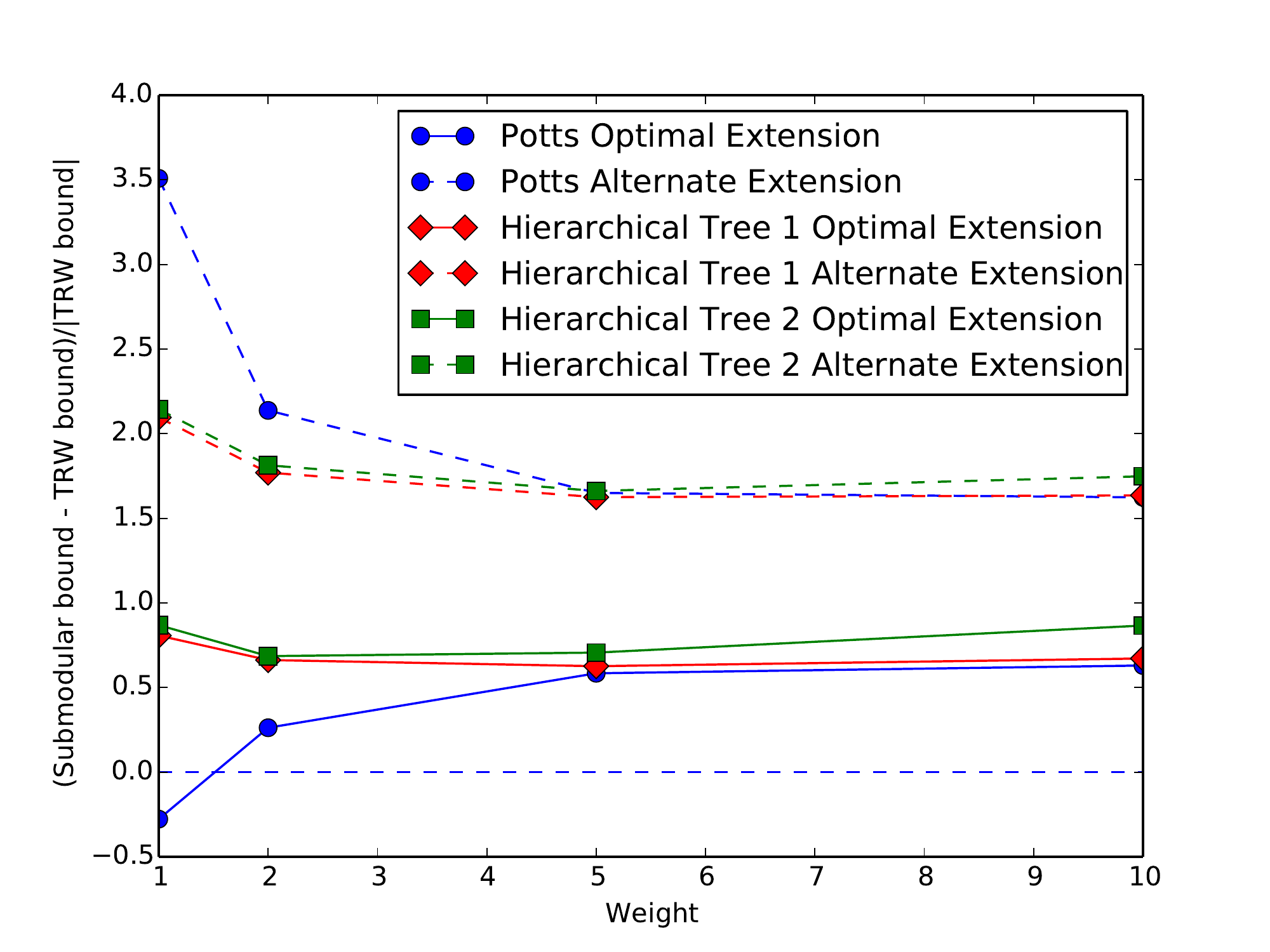}
\mycaption{\footnotesize \em Upper-bound comparison using synthetic data. The
plot shows the ratio (Submod bound - TRW bound)/|TRW bound| averaged over 100
unary instances as a function of pairwise weights using the worst-case optimal and
alternate extension for Potts and hierarchical Potts models. We observe that
the worst-case optimal extension
(solid) results in tighter bounds as compared to the respective alternate
extensions (dotted). Also, the worst-case optimal extension bounds are in
similar range as the TRW bounds. Figure best viewed in color.}
\label{fig:syn_plot}
\end{figure}

\mysubsection{Upper-bound Comparison using Synthetic Data}
\label{subsec:synthetic}
\myparagraph{Data} We generate lattices of size 100 $\times$ 100, where each
lattice point represents a variable taking one of 20 labels. The pairwise
relations of the sparse CRFs are defined by 4-connected neighbourhoods. The
unary potentials are uniformly sampled in the range [0, 10]. We consider (a)
Potts model and (b) hierarchical Potts models with pairwise distance between
labels given by the trees of figure \ref{fig:syn_rhst}. The pairwise weights
are varied in the range $\{1, 2, 5, 10\}$. We compare the results of
our worst-case optimal submodular extension with an alternate submodular
extension as given in figure \ref{fig:alternate}.

\myparagraph{Method} For our algorithm, we use the standard schedule $\gamma = 2/(k + 2)$ to obtain step size $\gamma$ at iteration $k$. We run our algorithm till convergence - 100 iterations suffice for this. The experiments are repeated for 100 randomly generated unaries for each model and each weight. For TRW, we used the MATLAB toolbox of \citep{domke2013learning}. The baseline code does not optimise over tree distributions. We varied the edge-appearance probability in trees over the range [0.1 - 0.5] and found 0.5 to give tightest upper bound.

\myparagraph{Results} We plot the ratio of the normalised difference of the upper
bound values of our method with TRW as a function of pairwise weights. The
ratios are averaged over 100 instances of unaries. Figure \ref{fig:syn_plot}
shows the plots for Potts and hierarchical Potts models for the worst-case
optimal and alternate extension. We find that the optimal extension (solid) results 
in tighter upper-bounds than the alternate extension (dotted) for both models.
To see the reason for this, we observe that the representation of the
submodular function using figure \ref{fig:alternate_extension} necessitates
that $\phi_a(i)$ be non-negative. This implies that $F(A)$ values are larger
for the worst-case optimal extension of figure \ref{fig:optimal_extension_graph} as compared to the
alternate extension. Hence the minimisation problem \ref{eq:min_upper_bound}
has the same objective function $g({\bf s})$ for both cases but the domain
$EP(F)$ of equation \eqref{eq:epf} is larger for the optimal extension, thereby
resulting in better minima.

Figure \ref{fig:syn_plot} also indicates that our algorithm with optimal extension provides similar range of upper bound as TRW, thereby providing empirical justification of our
method. Note that the TRW upper bound has to be tighter
than our method. This is because the TRW makes use of the standard LP
relaxation \citep{chekuri2004linear} which involves marginal variables for nodes as well as edges. On the other hand, our method makes use of the LP relaxation proposed by
\citet{kleinberg2002approximation} which involves marginal variables only for nodes. The standard LP
relaxation is tighter than Kleinberg-Tardos relaxation, and hence TRW results
in better approximation. However, TRW does not scale well with neighborhood size, thereby prohibiting its use in dense CRFs.

%
%
%
%
%
%
%
\begin{figure*}
    \centering
\begin{tabular}{cccccc}
  \includegraphics[scale = 0.15]{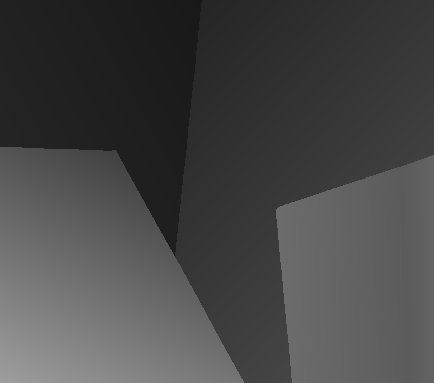} &
  \includegraphics[scale = 0.15]{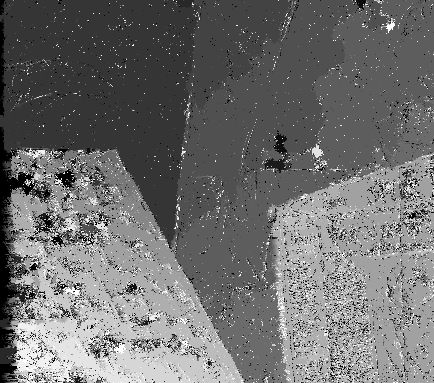} &
  \includegraphics[scale = 0.15]{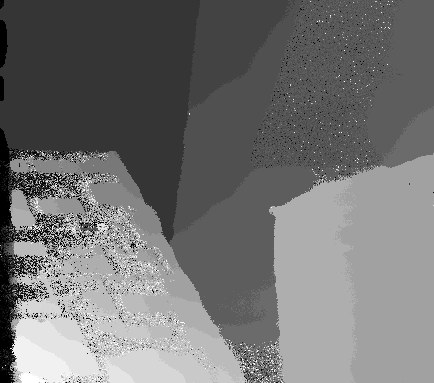} &
  \includegraphics[scale = 0.20]{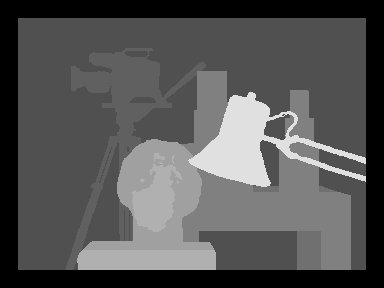} &
  \includegraphics[scale = 0.20]{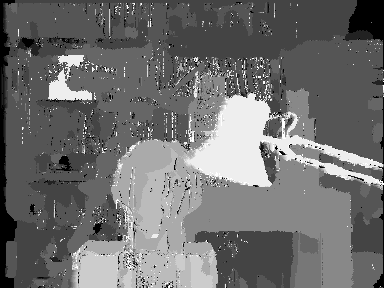} &
  \includegraphics[scale = 0.20]{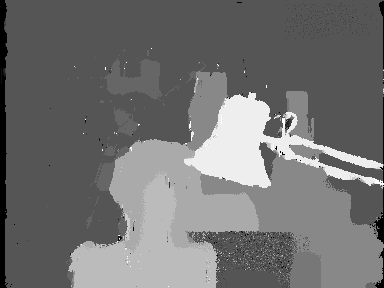} \\
        \scriptsize(a) Venus GT & \scriptsize(b) MF solution & \scriptsize(c) Submod solution & \scriptsize(a) Tsukuba GT & \scriptsize(b) MF solution & \scriptsize(c) Submod solution\\
        {} & \scriptsize  60.32s, 1.83e+07 &  \scriptsize 469.75s, {\bf 1.55e+07} & {} & \scriptsize  14.93s, 8.21e+06 & \scriptsize 215.22s, {\bf 4.12e+06} \\
        \includegraphics[scale = 0.15]{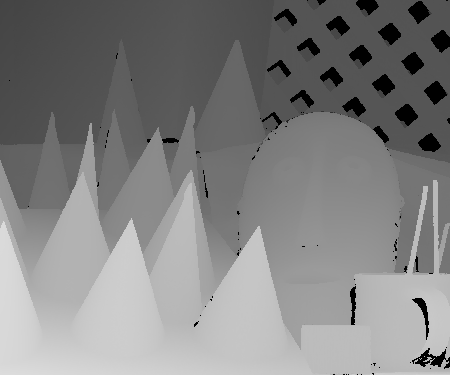} &
        \includegraphics[scale = 0.15]{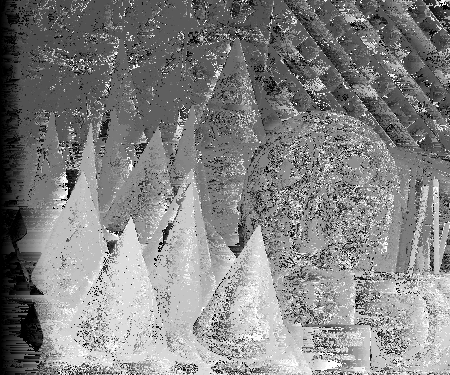} &
        \includegraphics[scale = 0.15]{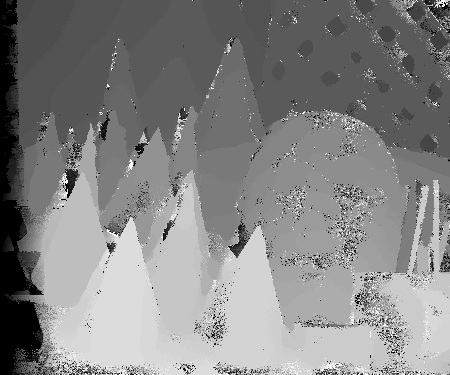} &
        \includegraphics[scale = 0.16]{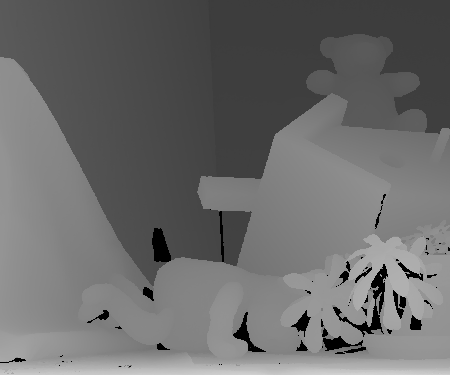} &
        \includegraphics[scale = 0.16]{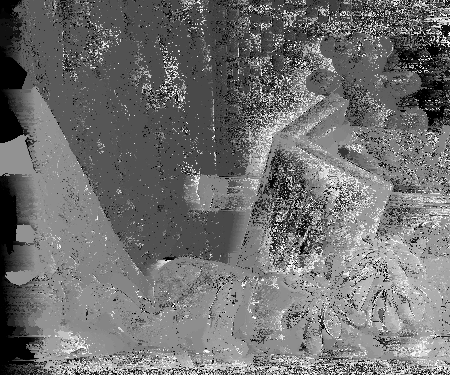} &
        \includegraphics[scale = 0.16]{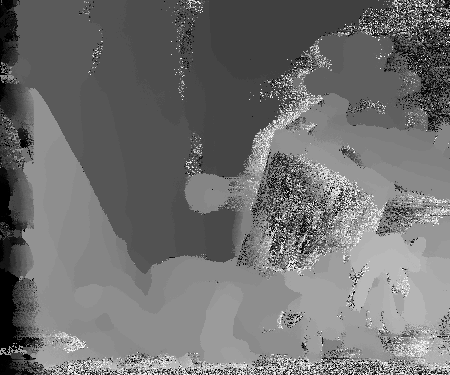} \\
    \scriptsize(a) Cones GT & \scriptsize(b) MF solution & \scriptsize(c) Submod solution & \scriptsize(a) Teddy GT & \scriptsize(b) MF solution & \scriptsize(c) Submod solution\\
        {} & \scriptsize 239.14s, 2.68e+07 &  \scriptsize 1082.72s, {\bf 1.27e+07} & {} & \scriptsize 555.30s, 2.36e+07 & \scriptsize 1257.86s, {\bf 1.58e+07}\\
\end{tabular}
\mycaption{\footnotesize \em Stereo matching using dense CRFs with Potts compatibility and Gaussian pairwise potentials. We compare our solution with the mean-field algorithm of \citet{koltun2011efficient}. We observe that our method gives better-looking solutions with lower energy value at the cost of higher computational time.}
\label{fig:stereo}
\end{figure*}

\mysubsection{Stereo Matching using Dense CRFs}
\label{subsec:stereo}
\myparagraph{Data} We demonstrate the benefit our algorithm for stereo matching on images extracted from the Middlebury stereo matching dataset \citep{scharstein2001taxonomy}. We use dense CRF models with Potts compatibility term and Gaussian pairwise potentials. The unary terms are obtained using the absolute difference matching function of \citep{scharstein2001taxonomy}. 

\myparagraph{Method} We use the implementation of mean-field algorithm for
dense CRFs of \citep{koltun2011efficient} as our baseline. For our algorithm,
we make use of the modified Gaussian filtering implementation for dense CRFs by
\citep{ajanthan2017efficient} to compute the conditional gradient at each step.
The step size $\gamma$ at each iteration is selected by doing line search. We
run our algorithm till 100 iterations, since the visual quality of the solution
does not show much improvement beyond this point. We run mean-field up to convergence, with a threshold of 0.001 for change in KL-divergence.

\myparagraph{Results}

Figure \ref{fig:stereo} shows some example solutions obtained by picking the
label with maximum marginal probability for each variable for mean-field and
for our algorithm. We also report the time and energy values of the solution
for both methods. Though we are not performing MAP estimation, energy values
give us a quantitative indication of the quality of solutions. For the full set of
21 image pairs (2006 dataset), the average ratio of the energies of the solutions from our method 
compared to mean-field is 0.943. The avearge time ratio is 10.66. We observe that our algorithm results in more natural looking stereo matching results with lower energy values for all images. However, mean-field runs faster than our method for each instance.

\mysection{Discussion}
\label{sec:conclude}
We have established the relation between submodular extension for the Potts
model and the LP relaxation for MAP estimation using Lovasz extension. This allowed us to identify the worst-case optimal submodular extension for Potts as well as the general metric labeling problems. It is worth noting that it might still be possible to obtain an improved submodular extension for a given problem instance. The design of a computationally feasible algorithm for this task is an interesting direction of future research. While our current work has focused on pairwise graphical models, it can be readily applied to high-order potentials by considering the corresponding LP relaxation objective as the Lovasz extension of a submodular extension. The identification of such extensions for popular high-order potentials such as the $P^n$ Potts model or its robust version could further improve the accuracy of important computer vision applications such as semantic segmentation.
\newpage

%% file: appendix_arxiv.tex
%
{\bf \LARGE Appendix}
\vspace{0.5cm}
\mysection{Proofs for Potts Model Extension} 

\paragraph{Remark 1} We show using induction over the number of variables that with $1$-of-$L$ encoding for Potts, 

\begin{equation}
    \sum_{A \in {\mathcal M}} \exp(-s(A)) = \prod_{a = 1}^{N} \sum_{i = 1}^L
    \exp(-s_{ai}).
    \label{eq:ind}
\end{equation}

\begin{proof}
Let $t$ be the number of variables, $V^t$ be the corresponding ground set and ${\mathcal M}^t$ be the sets corresponding to valid labelings. Equation \eqref{eq:ind} clearly holds for $t = 1$.

Let us assume that the relation holds for $t = N$, that is,

\begin{equation}
    \sum_{A^N \in {\mathcal M}^N} \exp(-s(A^N)) = \prod_{a = 1}^{N} \sum_{i = 1}^L \exp(-s_{ai})
\end{equation}

For $t = N + 1$, 

\begin{align}
    \sum_{A^{N + 1} \in {\mathcal M}^{N + 1}} \exp(-s(A^{N + 1})) &=  \sum_{i = 1}^L \sum_{A^N \in {\mathcal M}^{N}} \exp(-s(A^N) - s_{N+1, i}) \nonumber \\
                                                                  &= \sum_{i = 1}^L \exp(-s_{N+1, i}) \sum_{A^N \in {\mathcal M}^{N}} \exp(-s(A^N) ) \nonumber \\
                                                                  &= \sum_{i = 1}^L \exp(-s_{N+1, i}) \prod_{a = 1}^{N} \sum_{i = 1}^L \exp(-s_{ai}) \nonumber \\
                                                                  &= \prod_{a = 1}^{N + 1} \sum_{i = 1}^L \exp(-s_{ai})
\end{align}
\end{proof}

\paragraph{Remark 2} Given any submodular extension $F(.)$ of a Potts energy function $E(.)$, its Lovasz extension $f(.)$ defines an LP relaxation of the MAP problem for $E(.)$ as 
\begin{align}
    \min_{{\bf y} \in \Delta} f({\bf y}).
\end{align}
\label{lemma:lovasz_lp}
\begin{proof}
    By definition of a submodular extension and the Lovasz extension, $E({\bf x}) = F(A_{\bf x}) = f(1_{A_{\bf x}})$ for all valid labelings ${\bf x}$. Also, from property 1, $f({\bf y})$ is maximum of linear functions. Hence, $f({\bf y})$ is a piecewise linear relaxation of $E({\bf x})$.

    The domain $\Delta$ is a polytope formed by union of $N$ probability simplices
\begin{equation}
    \Delta = \{{\bf y}_a \in \mathbb{R}^L | {\bf y}_a \succeq 0  \textrm{ and } \langle {\bf 1}, {\bf y}_a \rangle = 1\}
    \label{eq:delta}
\end{equation}
With objective as maximum of linear functions and domain as a polytope, we have an LP relaxation of the corresponding MAP problem.
\end{proof}

{\proposition In the limit $T \to 0$, the following problem for Potts energies 
\begin{equation}
    \min_{{\bf s} \in EP(F)} \quad g_T ({\bf s}) = \sum_{a = 1}^{N} T \cdot \log \sum_{i = 1}^L \exp(-\frac{s_{ai}}{T}).
\label{eq:potts_temp}
\end{equation}

becomes
\begin{align}
  - \min_{{\bf y} \in \Delta} f({\bf y}).
\end{align}
\label{proposition:potts_equiv}}

\begin{proof}
In the limit of $T \to 0$, we can rewrite the above problem as
\begin{equation}
    \min_{{\bf s} \in EP(F)} \quad \sum_{a = 1}^{N} \max_{i} (-s_{ai}).
\end{equation}
In vector form, the problem becomes
\begin{align}
    &\min_{{\bf s} \in EP(F)} \max_{{\bf y} \in \Delta} - \langle {\bf y}, {\bf s} \rangle \\
    & = - \max_{{\bf s} \in EP(F)} \min_{{\bf y} \in \Delta} \langle {\bf y}, {\bf s} \rangle
    \label{eq:maxmin}
\end{align}
$\Delta$ is the union of $N$ probability simplices:
\begin{equation}
    \Delta = \{{\bf y}_a \in \mathbb{R}^L | {\bf y}_a \succeq 0  \textrm{ and } \langle {\bf 1}, {\bf y}_a \rangle = 1\}
    \label{eq:delta}
\end{equation}
where ${\bf y}_a$ is the component of $\bf y$ corresponding to the $a$-th variable. By the minimax theorem for LP, we can reorder the terms:
\begin{equation}
    - \min_{{\bf y} \in \Delta} \max_{{\bf s} \in EP(F)} \langle {\bf y}, {\bf s} \rangle 
    \label{eq:minmax}
\end{equation}
Recall that $\max_{{\bf s} \in EP(F)} \langle {\bf y}, {\bf s} \rangle$ is the value of the Lovasz extension of $F$ at $\bf y$, that is, $f({\bf y})$. Hence, as $T \to 0$, the marginal inference problem converts to minimising the Lovasz extension under the simplices constraint:
\begin{equation}
    - \min_{{\bf y} \in \Delta} f({\bf y}) 
    \label{eq:minLovasz}
\end{equation}
\end{proof}

{\proposition The objective function $E({\bf y})$ of the LP relaxation (P-LP) is the Lovasz extension of $F_{Potts}(A) = \sum_{i = 1}^L F_i(A)$, where
\begin{align}
    F_i(A) &= \sum_a \phi_{a}(i) [|A \cap \{v_{ai}\}| = 1] + \nonumber \\
           &\sum_{(a, b) \in {\mathcal N}} \frac{w_{ab}}{2} \cdot [|A \cap
           \{v_{ai}, v_{bi}\}| = 1].
\end{align}
\label{proposition:a-potts_optimal}}

\begin{proof}
Since $F_{Potts}$ is sum of Ising models $F_i$, we first focus on a particular label $i$ and then generalize. Consider a graph with only two variables $X_a$ and $X_b$ with an edge between them. The ground set in this case is $\{v_{ai}, v_{bi}\}$. Let the corresponding relaxed indicator variables be ${\bf y} = \{y_{aj}, y_{bj}\}$, such that $y_{ai}, y_{bi} \in [0, 1]$ and assume $y_{ai} > y_{bi}$. The Lovasz extension is:

\begin{align}
f({\bf y}) &= y_{ai} \cdot [F_i(\{v_{ai}\}) - F_i(\{\})] + y_{bi} \cdot [F_i(\{v_{ai}, v_{bi}\}) - F_i(\{v_{ai}\})] \nonumber \\
&= y_{ai} \cdot [\left(\phi_{a}\left(j\right) + \frac{w_{ab}}{2}\right) - 0] + y_{bi} \cdot [\left(\phi_{a}\left(j\right) + \phi_{b}\left(j\right)\right) - \left(\phi_{a}\left(j\right) + \frac{w_{ab}}{2}\right)] \nonumber \\
&=  \phi_{a}\left(j\right) \cdot y_{ai} + \phi_{b}\left(j\right) \cdot y_{bi} + \frac{w_{ab}}{2} \cdot \left(y_{ai} - y_{bi}\right)
\end{align}

In general for both orderings of $y_{ab}$ and $y_{bi}$, we can write
\begin{equation}
f({\bf y}) = \phi_{a}(j) \cdot y_{ai} + \phi_{b}(j) \cdot y_{bi} + \frac{w_{ab}}{2} \cdot |y_{ai} - y_{bi}|
\label{eq:a-lovasz_ising}
\end{equation}
Extending Lovasz extension (equation \eqref{eq:a-lovasz_ising}) to all variables and labels gives $E({\bf y})$ in (P-LP). 
\end{proof}

\mysection{Proofs for Hierarchical Potts Model Extension} 
\vspace{0.5cm}
\myparagraph{Transformed Tightest LP Relaxation} We take (T-LP) and rewrite it using indicator variables for all labels and meta-labels. Let $\mathcal R$ denote the set of all labels and meta-labels, that is, all nodes in the tree apart from the root. Also, let $\mathcal L$ denote the set of labels, that is, the leaves of the tree. Let $T_i$ denote the subtree which is rooted at the $i$-th node. We introduce an indicator variable $z_{ai} \in \{0, 1\}$, where
\begin{align}
    \enskip  z_{ai} =  \begin{cases} 
        y_{ai} \quad &\text{if} \enskip i \in {\mathcal L} \\
        y_{a}(T_i) \quad &\text{if} \enskip i \in {\mathcal R - \mathcal L} \\
    \end{cases}
\end{align}

We need to extend the definition of unary potentials to the expanded label space as follows:
\begin{align}
    \text{where} \enskip    \phi'_{a}(i) =  \begin{cases} 
        \phi_{a}(i) \quad &\text{if} \enskip i \in {\mathcal L} \\
        0  \quad &\text{if} \enskip i \in {\mathcal R - L} \\
    \end{cases}
\end{align}
We can now rewrite problem (T-LP) in terms of new indicator variables $z_{ai}$:
\begin{align}
    \text{(T-LP-FULL)} \quad &\min \widetilde{E}({\bf z}) = \sum_{i \in \mathcal{R}} \sum_{a \in {\mathcal X}} \phi_a'(i) \cdot z_{ai} + \nonumber \\
            &\sum_{i \in \mathcal{R}} \sum_{(a, b) \in {\mathcal N}} w_{ab} \cdot l_{T_i} \cdot |z_{ai} - z_{bi}| \nonumber \\
    \text{such that} \quad {\bf z} \in \Delta'
\label{eq:t-lp-full}
\end{align}
   where $\Delta'$ is the convex hull of the vectors satisfying
\begin{align}
    \enskip &\sum_{i \in \mathcal{L}} z_{ai} = 1, \enskip z_{ai} \in \{0, 1\} \enskip \forall a \in {\mathcal X}, i \in \mathcal{L} \\
    \text{and} \enskip &z_{ai} = \sum_{j \in L(T_i) } z_{aj}.\enskip \forall a \in {\mathcal X}, i \in \mathcal{R - L} \label{eq:consistency_constraint} 
\end{align}
Constraint (\ref{eq:consistency_constraint}) ensures consistency among labels and meta-labels, that is, if a label is assigned then all the meta-labels which lie on the path from the root to the label should be assigned as well. We are now going to identify a suitable set encoding and the worst-case optimal submodular extension using (T-LP-FULL).

\paragraph{Remark 3} Given any submodular extension $F(.)$ of a hierachical Potts energy function $E(.)$, its Lovasz extension defines an LP relaxation of the corresponding MAP estimation problem as
\begin{align}
  \min_{{\bf z} \in \Delta'} f({\bf z}).
\end{align}
\label{lemma:lovasz_lp_metric}

\begin{proof}
    By definition of a submodular extension and the Lovasz extension, $E({\bf x}) = F(A_{\bf x}) = f(1_{A_{\bf x}})$ for all valid labelings ${\bf x}$. Also, from property 1, $f({\bf y})$ is maximum of linear functions. Hence, $f({\bf y})$ is a piecewise linear relaxation of $E({\bf x})$.

    We can write the domain $\Delta'$ as 
    \begin{equation}
    \Delta' = \{ {\bf y}_a \in \mathbb{R}^M | {\bf y}_a \succeq 0, \enskip \langle {\bf 1}, {\bf y}_a^{label} \rangle = 1, \enskip {\bf y}_a(p_{ai}) = {\bf 1} \textrm{ or } {\bf y}_a(p_{ai}) = {\bf 0} \forall i \in [1, L]\}
    \label{eq:delta_tree_def}
    \end{equation} 
where ${\bf y}_a$ is the component of $\bf y$ corresponding to the $a$-th variable, ${\bf y}_a^{label}$  is the component of ${\bf y}_a$ corresponding to the $L$ labels, and ${\bf y}_a(p_{ai})$ is the component of ${\bf y}_a$ corresponding to the elements of $p_{ai}$. 

    Since $\Delta'$ is defined by linear equalities and inequalities, it is a polytope. With objective as maximum of linear functions and domain as a polytope, we have an LP relaxation of the corresponding MAP problem.
\end{proof}

{\proposition In the limit $T \to 0$, the following problem for hierarchical Potts energies 

\begin{equation}
\min_{{\bf s} \in EP(F)} \quad g_T ({\bf s}) = \sum_{a = 1}^{N} T \cdot \log \sum_{i = 1}^L \exp(-\frac{s'_{ai}}{T}).
\label{eq:metric_temp}
\end{equation}
becomes:
\begin{align}
  - \min_{{\bf z} \in \Delta'} f({\bf z}).
\end{align}
\label{proposition:metric_equiv}}

\begin{proof}
In the limit of $T \to 0$, we can rewrite the above problem as
\begin{equation}
    \min_{{\bf s} \in EP(F)} \quad \sum_{a = 1}^{N} \max_{i} (-s'_{ai}).
\end{equation}
In vector form, the problem becomes
\begin{align}
    &\min_{{\bf s} \in EP(F)} \max_{{\bf z} \in \Delta} - \langle {\bf z}, {\bf s'} \rangle \\
    & = - \max_{{\bf s} \in EP(F)} \min_{{\bf z} \in \Delta} \langle {\bf z}, {\bf s'} \rangle
    \label{eq:metric_maxmin}
\end{align}
\begin{equation}
    \text{where } \Delta = \{{\bf z}_a \in \mathbb{R}^L | {\bf z}_a \succeq 0  \textrm{ and } \langle {\bf 1}, {\bf z}_a \rangle = 1\}
    \label{eq:delta}
\end{equation}
where ${\bf z}_a$ is the component of $\bf z$ corresponding to the $a$-th variable. We can unpack $\bf s'$ using 
\begin{equation}
s'_{ai} = \sum_{t \in p_{ai}} s_t.
\label{eq:tree_s}
\end{equation}
and rewrite problem \eqref{eq:metric_maxmin} as
\begin{align}
    - \max_{{\bf s} \in EP(F)} \min_{{\bf y} \in \Delta'} \langle {\bf y}, {\bf s} \rangle
    \label{eq:metric_maxmin_unpack}
\end{align}
The new constraint set $\Delta'$ ensures that the binary entries of labels and meta-labels is consistent:
\begin{align}
    \textrm{where } \Delta' &= \{ {\bf y}_a \in \mathbb{R}^M | {\bf y}_a \succeq 0, \enskip \langle {\bf 1}, {\bf y}_a^{label} \rangle = 1, \nonumber \\
    &\enskip {\bf y}_a(p_{ai}) = {\bf 1} \textrm{ or } {\bf y}_a(p_{ai}) = {\bf 0} \forall i \in [1, L]\}
\label{eq:delta_tree_def}
\end{align} 
where ${\bf y}_a$ is the component of $\bf y$ corresponding to the $a$-th variable, ${\bf y}_a^{label}$  is the component of ${\bf y}_a$ corresponding to the $L$ labels, and ${\bf y}_a(p_{ai})$ is the component of ${\bf y}_a$ corresponding to the elements of $p_{ai}$. 

By the minimax theorem for LP, we can reorder the terms:
\begin{equation}
    - \min_{{\bf y} \in \Delta'} \max_{{\bf s} \in EP(F)} \langle {\bf y}, {\bf s} \rangle 
    \label{eq:metric_minmax}
\end{equation}
Recall that $\max_{{\bf s} \in EP(F)} \langle {\bf y}, {\bf s} \rangle$ is the value of the Lovasz extension of $F$ at $\bf y$, that is, $f({\bf y})$. Hence, as $T \to 0$, the marginal inference problem converts to minimising the Lovasz extension under the constraints $\Delta'$:
\begin{equation}
  - \min_{{\bf y} \in \Delta'} f({\bf y}).
    \label{eq:metric_minLovasz}
\end{equation}
\end{proof}

{\proposition The objective function $\widetilde{E}({\bf z})$ of (T-LP-FULL) is the Lovasz extension of $F_{r-\textrm{HST}}(A) = \sum_{i = 1}^M F_i(A)$, where
\begin{align}
    F_i(A) &= \sum_a \phi'_{a}(i) [|A \cap \{v_{ai}\}| = 1] + \nonumber \\
           &\sum_{(a, b) \in {\mathcal N}} {w_{ab}} \cdot l_{T_i} \cdot [|A
           \cap \{v_{ai}, v_{bi}\}| = 1].
\end{align}
\label{proposition:rhst_optimal}}
\begin{proof}
    We observe that $F_{r-\textrm{HST}}$ is of exactly the same form as $F_{Potts}$, except that the Ising models $F_i$ are defined over not just labels, but meta-labels as well. Using the same logic as in the proof of proposition \ref{proposition:a-potts_optimal}, each $F_i$ is the Lovasz extension of 
        \begin{equation}
            \widetilde{E}_i({\bf z}) = \left( \sum_{a \in {\mathcal X}} \phi_a'(i) \cdot z_{ai} + \sum_{(a, b) \in {\mathcal N}} w_{ab} \cdot l_{T_i} \cdot |z_{ai} - z_{bi}| \right)
        \end{equation}
    and the results follows. 
\end{proof}